\documentclass[sigconf]{acmart}

\usepackage{subcaption}
\usepackage{enumitem}

\usepackage{array}
\newcolumntype{P}[1]{>{\raggedright\arraybackslash}p{#1}}
\usepackage{multirow}

\AtBeginDocument{%
  \providecommand\BibTeX{{%
    \normalfont B\kern-0.5em{\scshape i\kern-0.25em b}\kern-0.8em\TeX}}}

\setcopyright{none}

\copyrightyear{2023} 
\acmYear{2023} 
\setcopyright{rightsretained} 
\acmConference[FAccT '23]{2023 ACM Conference on Fairness, Accountability, and Transparency}{June 12--15, 2023}{Chicago, IL, USA}
\acmBooktitle{2023 ACM Conference on Fairness, Accountability, and Transparency (FAccT '23), June 12--15, 2023, Chicago, IL, USA}
\acmDOI{10.1145/3593013.3594089}
\acmISBN{979-8-4007-0192-4/23/06}

\begin{document}

\title{Organizational Governance of Emerging Technologies: \\AI Adoption in Healthcare}

\author{Jee Young Kim}
\affiliation{%
  \institution{Duke Institute for Health Innovation}
  \country{United States}
}
\author{William Boag}
\affiliation{%
  \institution{Duke Institute for Health Innovation}
  \country{United States}
}
\author{Freya Gulamali}
\affiliation{%
  \institution{Duke Institute for Health Innovation}
  \country{United States}
}
\author{Alifia Hasan}
\affiliation{%
  \institution{Duke Institute for Health Innovation}
  \country{United States}
}
\author{Henry David Jeffry Hogg}
\affiliation{%
  \institution{Newcastle University}
  \country{United Kingdom}
}
\author{Mark Lifson}
\affiliation{%
  \institution{Mayo Clinic}
  \country{United States}
}
\author{Deirdre Mulligan}
\affiliation{%
  \institution{University of California, Berkeley}
  \country{United States}
}
\author{Manesh Patel}
\affiliation{%
  \institution{Duke University Medical Center}
  \country{United States}
}
\author{Inioluwa Deborah Raji}
\affiliation{%
  \institution{University of California, Berkeley}
  \country{United States}
}
\author{Ajai Sehgal}
\affiliation{%
  \institution{Mayo Clinic}
  \country{United States}
}
\author{Keo Shaw}
\affiliation{%
  \institution{DLA Piper}
  \country{United States}
}
\author{Danny Tobey}
\affiliation{%
  \institution{DLA Piper}
  \country{United States}
}
\author{Alexandra Valladares}
\affiliation{%
  \institution{Community Representative}
  \country{United States}
}
\author{David Vidal}
\affiliation{%
  \institution{Mayo Clinic}
  \country{United States}
}
\author{Suresh Balu}
\affiliation{%
  \institution{Duke Institute for Health Innovation}
  \country{United States}
}
\author{Mark Sendak}
\affiliation{%
  \institution{Duke Institute for Health Innovation}
  \country{United States}
}

\renewcommand{\shortauthors}{Kim et al. 2023}

\begin{abstract}

Private and public sector structures and norms refine how emerging technology is used in practice. In healthcare, despite a proliferation of AI adoption, the organizational governance (i.e. institutional governance) surrounding its use and integration is often poorly understood. 
What the Health AI Partnership (HAIP) aims to do in this research is to better define the requirements for adequate organizational governance of AI systems in healthcare settings and support health system leaders to make more informed decisions around AI adoption. To work towards this understanding, we first identify how the standards for the AI adoption in healthcare may be designed to be used easily and efficiently. Then, we map out the precise decision points involved in the practical institutional adoption of AI technology within specific health systems. 
Practically, we achieve this through a multi-organizational collaboration with leaders from major health systems across the United States and key informants from related fields. Working with the consultancy IDEO.org, we were able to conduct usability-testing sessions with healthcare and AI ethics professionals. Usability analysis revealed a prototype structured around mock key decision points that align with how organizational leaders approach technology adoption. Concurrently, we conducted semi-structured interviews with 89 professionals in healthcare and other relevant fields. Using a modified grounded theory approach, we were able to identify 8 key decision points and comprehensive procedures throughout the AI adoption lifecycle. 
This is one of the most detailed qualitative analyses to date of the current governance structures and processes involved in AI adoption by health systems in the United States. We hope these findings can inform future efforts to build capabilities to promote the safe, effective, and responsible adoption of emerging technologies in healthcare.

\end{abstract}



\sloppy
\maketitle

\section{Introduction}

There is a long acknowledged need to think constructively and proactively about the assets and deficits of distinct public and private actors that can be harnessed to meet governance objectives \cite{freeman2000private}. 
Examining how powerful entities in a domain are approaching the governance of algorithmic 
decision-making provides insight into current capabilities and competencies, as well as factors limiting the emergence of sound governance models. Studying private governance activities allows us to assess their alignment with the public values that guide the health care field, and identify opportunities to ``catalyze the ongoing development of meaningful internal practices,'' \cite{bamberger2011new} aligning with recent calls for ``studying up'' into the details of institutional decision-making as a means of holding powerful actors  accountable \cite{barabas2020studying}.

Our project documents the approaches individual health care systems are adopting to govern the use of AI in clinical care.  
Healthcare is a particularly compelling case study because the domain-specific understanding of quality encompasses equity and ethics. Additionally, a wide range of entities contribute to health care governance in various verticals providing policymakers a useful example of the range of entities, with different core competencies and capacities that can be enlisted to successfully produce quality healthcare as AI is introduced into clinical care at an increasing rate. 

The current study describes early stage research undertaken to support a multi-organizational effort to promote the safe, effective, and equitable adoption of AI software by health systems in the United States context specifically. Our findings are relevant to organizational leaders looking to improve internal algorithm adoption processes, algorithm developers who sell to large, regulated enterprises, and policy makers looking to improve regulation of emerging technologies. Our contributions are as follows: first, we identify urgent needs involved in establishing governance structures for the use of AI technology in healthcare organizations and define usable resources to support health system leaders; second, we examine the current state of practical institutional adoption of AI technology within specific health systems and surface the precise decision points. 


\section{Background}
\label{sec:background}

Various verticals of healthcare technology–such as pharmaceutical therapeutics, implanted devices, laboratory diagnostics–already feature mature ecosystems of affected parties that collaboratively ensure safe and effective use of new technologies in clinical care. Each vertical is subject to their own governance standards from regulators, health insurance payers, professional licensing bodies, and accreditation bodies. The new technologies are each governed through dedicated organizational structures within health systems. For example, laboratory quality management systems oversee testing, and value analysis teams oversee devices. These organizational governance structures often oversee adoption of both internally and externally developed technologies and are staffed by personnel with specialized training in the technology vertical. Where AI is embedded in healthcare technology within these verticals, governance processes may need new tools and expertise. However, much of the AI in development falls outside these domains. 
 
There is currently no clear and established governance ecosystem specifically for AI use in healthcare. In fact, many algorithms used for clinical decision support or for prioritizing access to specialized services may be outside the legislative governance purview of the US Food and Drug Administration (FDA). Payment models (and the incentives they encode) for healthcare algorithms are emerging, but almost exclusively apply to algorithms regulated by FDA as devices \cite{parikh2022paying}. Information asymmetries undermine quality assurance within the healthcare AI market by preventing potential health system purchasers from understanding algorithm development and performance over time and across settings. A dearth of inter or multi disciplinary trained experts exacerbate this problem leaving all but the largest research and teaching healthcare systems ill equipped to develop AI systems internally or evaluate those offered by third parties. Health systems are increasingly called upon to enhance organizational governance of algorithms, but concrete guidance and capacity building programs are largely missing \cite{eaneff2020case}. And even opportunities for individual clinicians to upskill and participate directly in algorithm development or validation are limited \cite{otlecs2022teaching}. Drawing on the Swiss Cheese Model of Pandemic Defense \cite{nyt2018:maven}, every layer of the healthcare AI ecosystem currently contains large holes that make the broad diffusion of poorly performing products inevitable.

\subsection{Organizational Governance of Healthcare Technologies}

Healthcare is both highly regulated and rapidly evolving. In 2022 alone, the FDA approved 37 new medications \cite{us2022advancing} and approved or cleared thousands of devices \cite{us2022artificial}. Some healthcare uses of AI fall outside of FDA’s purview. This is generally because certain software is excluded from the definition of medical device \cite{food2020drug} and certain uses of medical AI may constitute the practice of medicine, which FDA does not regulate \cite{aboutfda}. FDA has historically exercised generous enforcement discretion for many diagnostic tests \cite{congressionalresearchservices}. Irrespective of FDA regulatory status, health systems must ensure the safety, efficacy, and equity of new innovations.
Below, we detail the affected parties and processes that support organizational governance of medications, tests, interventions, and devices towards this end.

Health systems that develop medications and tests for internal use are supported by additional actors, including standards setting bodies and third-party accreditors. For example, health care organizations often provide pharmacy compounding services, which refers to the practice of a licensed pharmacist combining, mixing, or altering ingredients of a drug to create a medication tailored to the needs of an individual patient \cite{humandrugcompounding}. Although there are many well-documented compounding errors that harmed patients \cite{watson2021pharmaceutical}, legislative governance falls short. The FDA inspects a limited scope of compounding pharmacies. To fill this gap, the US Pharmacopeia (USP) develops standards for medication compounding as well as accompanying training for pharmacists \cite{uspcompounding}. Third party accreditors, such as the Joint Commission, offer certification programs that ensure compounding pharmacies adhere to USP standards \cite{mdccertification}. 
 
Similar actors perform quality control functions for laboratory tests. As mentioned above, while some diagnostic tests are subject to FDA approval, laboratory developed tests used within individual facilities are not. The Centers for Medicare and Medicaid Services (CMS) determines payment to health systems for both FDA approved tests as well as laboratory developed tests. To complement FDA approval of individual tests, CMS regulates laboratory testing facilities through Clinical Laboratory Improvement Amendments (CLIA) certification. Only CLIA certified laboratories can receive Medicare or Medicaid payments for healthcare. Professional societies, such as the College of American Pathologists, and the Centers for Disease Control and Prevention (CDC) provide resources and guidance to support laboratory-directed quality management programs. The Clinical and Laboratory Standards Institute (CLSI) also develops international standards for clinical laboratory testing and offers extensive training for laboratory leaders. Lastly, numerous accredited fellowships train clinical pathologists to lead various laboratory testing activities \cite{acgme}.

Organizational governance of devices prominently features group purchasing organizations. Like medications, the FDA first reviews safety and efficacy data to approve individual devices for clinical use. CMS then determines payment for FDA-approved devices. Once CMS reimbursement and FDA approval are in place, adoption of the device by health systems and individual clinicians picks up. Health systems rely on internal value analysis teams or committees to evaluate medical device requests from clinicians and maintain device formularies \cite{engelman2018addressing}. Once a device is approved for the formulary, health systems work with external group purchasing organizations to negotiate rates with the manufacturer and supply the device \cite{bruhn2018group}. The group purchasing organization often conducts its own analyses and benchmarking and leverages buying power to prioritize high quality devices for health systems. 

\subsection{Cautionary Tale: Antimicrobial Stewardship}
Complex networks of entities inside and outside health systems support organizational governance of emerging technologies used in clinical care. However, their “Swiss Cheese” structure has limitations, creating a failure of organizational governance. It took nearly 75 years from the time Alexander Fleming identified the harms of antibiotic resistance for affected parties and regulators to enact governance structures (Appendix \ref{appendix:antimicrobial}). During this time, the damage caused by antibiotic resistance grew to overshadow the harms of individual pathogens. In 2019 alone, there were 1.27 million deaths worldwide attributable to antibiotic resistance \cite{collaborators2022articles}, greater than the death toll due to either HIV or malaria. Antimicrobial stewardship programs provide a model for nascent efforts at "algorithmic stewardship," but the damage that occurred due to their after-the-fact adoption offers a compelling cautionary tale counseling proactive attention to algorithmic governance \cite{eaneff2020case}. Today, health systems are managing the downstream consequences of a collective failure to govern the introduction of antibiotics. Building an ecosystem to ensure sound algorithmic stewardship–both ensuring they improve quality and managing emerging risks–today can help us avert a similar fate.

\section{Recruiting Collaborators}
In April 2022, we launched the Health AI Partnership (HAIP) and recruited 11 healthcare organizations and 2 ecosystem partner sites in the United States. The goal of this effort was to empower healthcare professionals to use AI effectively, safely, and equitably through community-informed contemporary standards. The following types of organizations were included in this effort: 6 large academic health systems (Duke Health, Hackensack Meridian Health, Jefferson Health, Michigan Medicine, NewYork-Presbyterian, UCSF Health), 2 large health systems with locations across multiple US regions (Kaiser Permanente and Mayo Clinic), 2 non-profit health technology vendors that support safety net hospitals and community health centers (OCHIN and Parkland Center for Clinical Innovation), 1 large Medicaid health plan (WellCare NC), and 2 ecosystem partner sites (DLA Piper and UC Berkeley).


\section{Designing AI Adoption Standards}

Between July and August 2022, we conducted a 6-week design sprint with a team of designers from IDEO.org. The goal of this collaboration was to 
guide future development of products and programs that would build capabilities in health systems for making informed decisions around adopting emerging technologies. We wanted to design products and programs that health system leaders found immediately useful in practice. Through this design sprint, we wished to engage teams from across sites to identify how the products and programs might be most impactful. This research was considered a quality improvement project without the involvement of human subjects. Thus, it did not require an IRB approval. Leaders from across sites consented to participate in the design sprint.


Participants were recruited from the partner sites. Primary data were collected through 1 usability-testing session with project leaders, 3 usability-testing sessions with health system leaders, 2 usability-testing sessions with external AI ethics experts, and 1 usability testing session with representatives from the partner sites.

All sessions were conducted via Zoom for 1-2 hours in groups of 15-30 people composed of participants and IDEO.org designers. The sessions were unscripted and contained various activities that required active engagement of the participants. The activities were designed to understand (1) where an AI implementation would make the most sense, (2) who would be the decision makers in the implementation process, (3) how the decision makers would learn about new technologies and be motivated to implement AI, and (4) what roles ethics, equity, and justice might play into the implementation process.

IDEO.org synthesized experiences and needs that the participants expressed during the usability-testing sessions and drafted initial insights and paper prototypes (i.e. sample drafts of a final product). The initial insights and prototypes were iterated upon with the project team. We then shared the insights and prototypes with the health system leaders to incorporate their feedback and ensure that their experiences and needs were accurately captured.


Major insights highlighted opportunities for future work (Appendix \ref{appendix:ideo-insights}). These insights surfaced the urgent needs for a shared and objective guideline for adoption of AI solutions and community among health system leaders. More importantly, a simple prototype was created for the shared and objective AI adoption guideline (Appendix \ref{appendix:ideo-insights}). The prototype was structured around mock key decision points that align with how organizational leaders approach technology adoption. The design insights and prototype provided scaffolding for our team to build upon with more rigorous and comprehensive qualitative research.

\section{Surfacing Practical Decision Points}

Between April 2022 and January 2023, we conducted qualitative research using in-depth interviews, concurrent with design research. The goal of this work was to understand the current and aspirational state of AI adoption in healthcare organizations. We wished to identify requirements for the adoption of AI systems in healthcare settings and map out adequate organizational governance structures. The design research informed how qualitative research results are framed around key decision points. We combined design research and qualitative research approaches to ensure that future product and program development addresses real needs of health system leaders. This research was considered a quality improvement project without the involvement of human subjects. Thus, it was exempted from an IRB approval. All participants provided verbal consent to be interviewed and to have anonymized data used in qualitative analyses.

\subsection{Method}

First, we developed an interview guide and identified which types of internal and external roles to interview based on four relatively distinct stages of AI adoption: (1) problem identification and procurement, (2) development and adaptation, (3) clinical integration, and (4) lifecycle management \cite{sendak2020path} (Table \ref{tab:four-stages} in Appendix \ref{appendix:four-stages}). Interview guides were designed to walk participants through these four stages. Questions in the interview guide covered processes and personnel involved in each stage (examples in Appendix \ref{appendix:four-stages}).

Then, 89 professionals in healthcare and other relevant fields were recruited to participate in interviews. We used purposive and snowball sampling approaches to recruit participants. Participation was completely voluntary. All participants whom we reached out to agreed to participate in the study. Of the 89 participants, 70 were from the partner sites. 

Among the 70 partner site participants, there was diverse representation across clinical (n=12), technical (n=34), operational (n=19), and regulatory (n=5) roles. Frontline clinicians who were users or champions of AI systems were considered clinical personnel. Technical personnel were those who develop technology and IT infrastructure, including data scientists, data engineers, and IT leaders. Operational personnel were those in charge of managing operations within a health system department or facility. Regulatory personnel were those who ensure compliance with regulations. Two of the technical personnel and two of the operational personnel had joint appointments with clinical roles. 

Of the 89 participants, 19 were key informants with primary affiliations outside of healthcare organizations. Key informants were identified based on expertise in areas critical to safe and responsible adoption of AI software, including bias (n=10), ethics (n=3), community engagement (n=3), organizational behavior (n=1), regulation (n=1), and credentialing (n=1). Key informants were recruited to capture perspectives that were poorly represented by health system participants. 

All interviews were conducted via Zoom by 1-4 project leaders. All interviews were conducted with a single interviewee, except for two sessions where 2 and 4 participants from the same organization participated in each. Each interview was about an hour long, ranging from 34 to 82 minutes, and a timer was used to spend at least 10 minutes on each stage. The participants were asked the questions described in the interview guide, with a semi-structured approach to capture participants’ varying scope of experience and emergent concepts. The interviews were recorded once the participants gave a verbal consent and later were transcribed for analysis via Otter.ai, an automated transcription software. All participants gave informed consent for the interview and recording. We analyzed data alongside the recruitment and decided to terminate data collection when no new themes or insights emerged (i.e. thematic saturation). 

To analyze the transcripts, we followed a modified grounded theory \cite{charmaz2014constructing} approach and practiced a coding process to organize qualitative data. In our study, coding was conducted in 2 cycles. In the first cycle, we used descriptive open coding to develop initial codes and capture the general ideas that emerged from the raw data. In the second cycle, we used focused coding to examine the most frequent and significant codes and categorize the initials codes based on thematic similarity. After the second cycle of coding, we identified the most prevalent themes and subthemes, using the key decision point framework designed in Part 1. Data were analyzed using NVivo Qualitative Data Analysis Software.

\subsection{Results}

Major themes emerged as important factors that shaped decisions related to adopting AI software (Table \ref{tab:decision-points}). The factors were grouped into 8 thematic areas that described key decision points across the entire process of adopting AI. Under each thematic area, subthemes were identified with representative quotations that captured challenges and corresponding best practices. The results identified  challenges and risks that organizations experience in the process of AI adoption as well as real-world processes and practices.

\begin{table*}
\label{tab:decision-points-overall}
  \begin{center}
    \caption{Key decision points represented as thematic areas and corresponding subthemes.}
    \label{tab:decision-points}
    \begin{tabular}{|m{10cm}|m{6cm}|}
        \hline
        Thematic Area & Subtheme \\ \hline
1. Identify and prioritize a problem &  
\begin{itemize}[noitemsep]
    \item Problem identification
    \item Problem prioritization
    \item AI contribution
\end{itemize}
\\ \hline
2. Identify requirements for an AI product as a viable component of the solution &  
\begin{itemize}
    \item Requirement and feasibility
    \item Quality assurance
    \item Build vs. buy decision
    \item Aligning affected parties
\end{itemize}
 \\ \hline
3. Develop measures of outcomes and success of the AI product &  
\begin{itemize}
    \item Model performance measures
    \item Success measures
\end{itemize}
\\ \hline
4. Design a new optimal workflow to facilitate integration &  
\begin{itemize}
    \item Operational optimization
    \item Technical optimization
\end{itemize}
\\ \hline
5. Evaluate safety, effectiveness, and equity concerns of the AI product in the intended setting prior to clinical use
 &  
\begin{itemize}
    \item Data quality assurance
    \item Validation
    \item Risk mitigation
    \item Clinically integrate vs. abandon decision
\end{itemize}
\\ \hline
6. Execute AI product rollout, workflow integration, communication, education, and scaling &  
\begin{itemize}
    \item Communication
    \item Education and training
    \item Trust
\end{itemize}
\\ \hline
7. After operationalization, monitor and maintain the AI product and impacted ecosystem &  
\begin{itemize}
    \item Technical monitoring
    \item Operational monitoring
    \item Outcome monitoring
    \item Cadence of monitoring reviews
    \item Accountability and ownership
\end{itemize}
\\ \hline
8. Update or decommission the AI product and impacted ecosystem &  
\begin{itemize}
    \item Update
    \item Decommission
\end{itemize}
\\ \hline
    \end{tabular}
  \end{center}
\end{table*}

\subsection{Identify and prioritize a problem}
This thematic area describes how the organizations identify a high-priority problem and consider adopting AI software to solve the problem. Representative quotations are presented in Appendix \ref{appendix:quotes-1}. 

\subsubsection{Problem identification}
The participants reported two dominant ways problems were initially identified as meriting investment. In some organizations, frontline workers surfaced daily pain points and brought them to organizational leaders for consideration. In others, problem identification was primarily led by organizational leadership, even before a need was escalated from frontline workers. Organizational leaders perceived the problems as emerging opportunities for the organizations to grow and identified areas in which the organizations might benefit from investments. Some organizations reported a mixed approach. The participants described the strengths and limitations of the two approaches. 

The participants highlighted the importance of understanding the context in which the problems were identified, especially when the organizational leaders raised the problems. To characterize the problems effectively, it was important to include frontline workers to provide an authentic sense of context. Involving members of underserved communities from the early stage of problem identification was raised as an aspirational and impactful practice. Once problems were identified, interviewees described the importance of defining the end users for the potential solution. In this process, it was important to understand that end users of the potential solution might not always be the ones who would benefit directly from implementing the solution. Sometimes, other parts of the organizations or downstream users could receive direct benefits from the work of end users. Given that the needs of end users and beneficiaries of the solution might not align, interviewees emphasized the  importance of engaging end users in the early stage of problem formulation to anticipate and mitigate these potential barriers to adoption.
 
The participants also noted that the organizations were frequently approached by vendors, and there could be external pressures to procure AI software. For instance, a clinical leader with a prior connection to a vendor could initiate the procurement of AI software without a thorough, centralized evaluation. The participants cautioned that such external forces might distract problem-led procurement which emerged as a best practice. In response to this possible risk, we observed that some organizations founded governance committees with enterprise-wide scope to prevent hasty procurement happening in silos.

\subsubsection{Problem prioritization}

Once the problems were identified, in some organizations, they were prioritized through a centralized process based on organizational criteria. In others, business units within the organization were allowed to make their own decisions guided by organizational criteria applied to their specialty. Criteria varied across organizations, including feasibility of solving the problems, severity of clinical condition, number of patients impacted by it, sense of urgency, use of pre-existing digital infrastructure, organizational goals and strategic initiatives. The participants commonly expressed a need to improve patient care and promote work efficiency by lessening the burden on clinicians.
 
The participants flagged the unequal distribution of resources within many organizations. Some departments or organizational business units might benefit from greater resources than others and thus have greater capabilities to identify emerging opportunities and solve problems. In that case, problems experienced by groups with greater resources, such as departments of radiology or cardiology, could be prioritized over problems of the groups with fewer resources, such as departments of pediatrics or women’s health. The participants suggested that organizations should allocate resources in an equitable fashion that does not further entrench existing disparities across business units.

\subsubsection{AI contribution}
When participants were asked how AI could be optimally used to solve problems, responses and proposed use cases varied tremendously. Yet, a commonly described unique contribution of AI was the use of data to promote efficiency in clinical care. For example, interviewees described the potential for AI  to help triage patients with different levels of risk, allocating physician time efficiently, and making informed decisions quickly.
 
Despite the potential benefits of using AI, participants emphasized the importance of well defined, narrow intended uses of AI. It was felt important to remember that AI software is there to help clinicians make better use of their time, not to replace clinicians or their work. The participants also reported that it is important to verify whether AI is indeed an appropriate solution to a problem. Participants described many instances in which software with simple decision logic can be a viable solution with greater transparency and lower resource requirements than AI or machine learning.

\subsection{Identify requirements for an AI product as a viable component of the solution}

This thematic area describes how organizations assess the feasibility and viability of adopting AI software to solve a prioritized problem. During this initial evaluation process, the organizations identify required resources, assess the feasibility of adoption, conduct preliminary quality assurance of the AI software, seek alignment among affected parties, and determine whether to build or buy the AI software. Representative quotations are presented in Appendix \ref{appendix:quotes-2}.

\subsubsection{Requirement and feasibility}
The resources identified by participants as pre-requisites to adopting AI software were extensive. The resources included funding, time, data environment, IT infrastructure, regulatory approval, and employees with skills and expertise in data science, clinical practice, data engineering, and software engineering. The participants noted that complying with regulation would require additional resources.
Assessing requirements and feasibility was an important step for determining whether AI would be a viable component of a solution. Yet, the participants described how these assessments were often inefficient, with resources and capabilities decentralized and duplicated across the organization. To enhance efficiencies, participants recommended centralizing roles and capabilities to think broadly about the optimal use of AI within the organization.

\subsubsection{Quality assurance}
Participants reported ways that they ensured the quality of AI software under consideration, especially when they decided to buy AI software from external vendors. They conducted literature reviews. 
They examined whether the AI software had an accurate and equal performance across different subgroups of patients, whether it had been validated for their patient population, and whether it was built on valid data sets that were easily accessible. They also inspected whether the AI software could be easily integrated in the existing clinical workflow and IT interface, whether it complied with regulation, and whether it could be sustainably used and expanded for other use cases. 
In this process, the participants reported difficulty with accessing the necessary material, unless the AI software was built in house. Vendors often did not have or were not willing to share the material. Given the lack of documentation and the amount of due diligence required, the participants raised a need for a standardized structure of quality assurance and active communication across organizations.

\subsubsection{Build vs. buy decision}
Participants felt motivated to build AI in-house when it did not exist in the market or they were skeptical about an existing model. 
They had multiple concerns about an external AI, including that it could be inappropriate for their target problem or patient population or that it could impose greater risk tolerance for meeting regulatory compliance. 
They highlighted a lack of agreed upon practices for model development and validation. A primary benefit of building AI internally was having full transparency \& visibility into the model development process. 
On the other hand, when the valid AI software already existed in the market or the organizations lacked the resources to build, including data for validation and internal expertise, the participants felt motivated to buy AI from external vendors. The benefit of buying was that the organizations could focus resources saved by not developing the AI on its implementation, which was seen as the area where organizations had the most to learn. Yet, it did not mean that buying AI software was necessarily cheaper than building one. In fact, buying AI software was sometimes seen as more expensive, as adapting the model to a setting where it would be integrated was costly. Also, buying AI software did not exempt the organizations from complying with regulations. Regardless of whether the organizations decided to build or buy AI software, they all felt the burden of potential legal and liability risks.

\subsubsection{Aligning affected parties}
The participants reported that in making decisions to develop or adapt AI software, all affected parties from various disciplines, including AI specific committees and specialists in clinical, operational, and technical areas, must be aligned. Challenges in aligning affected parties included identifying key affected parties who should be involved in making the decision and aligning different priorities across all affected parties. 

\subsection{Develop measures of outcomes and success of the AI product}

This thematic area describes how the organizations set their goals and defined measurable outcomes. Representative quotations are presented in Appendix \ref{appendix:quotes-3}.

\subsubsection{Model performance measure}

Sensitivity, specificity, area under the receiver operating characteristic curve, positive predictive value number, and false positive and negative rate were commonly referenced by interviewees as model performance metrics. Some of these were felt to map more intuitively to a given use case or end user experience than others. In defining the model metrics, it was seen to be important to ensure that the model performs similarly on subgroups stratified by important demographic factors so that patients receive equitable care.

\subsubsection{Success measure}
In addition to the model performance metrics, participants felt it was important to define which metrics the organizations should measure to determine the success of the solution. Outcomes commonly captured in the success metrics included improvements in patient outcomes, reductions in burden on healthcare providers, and reductions in cost. A challenge to developing success measures was reconciling the different definitions of success across different affected parties.

\subsection{Design a new optimal workflow to facilitate integration}

This thematic area describes how organizations develop workflows and user experiences surrounding AI software to facilitate its adoption into clinical care. Representative quotations are presented in Appendix \ref{appendix:quotes-4}.

\subsubsection{Operational optimization}
The participants reported that a new workflow involving AI software should be designed before the software was integrated into clinical practice. As algorithm developers tend to have limited insight into an existing workflow, it was critical that end users of the AI software were involved in the process of developing the new workflow and provided context into the current state. The participants proposed to design the new workflow in line with the existing one to minimize the changes made in implementation. Given that implementation of the AI software would create inevitable changes in the existing workflow, they also highlighted the importance of recognizing the changes and providing necessary support to front-line workers reconciling these changes. 

\subsubsection{Technical optimization}

For the AI software to cause minimal friction, designing an intuitive user interface (UI) was also an important component for optimizing technical aspects of workflow. One of the most common practices of building a user-friendly UI was to make it simple and customize it for different groups of users and different sites of users. Simplicity in the user interface could be achieved by standardizing the visualization of the model, conducting usability tests with users, and using technology that was already familiar to users.

\subsection{Evaluate safety, effectiveness, and equity concerns in the intended setting prior to clinical use}

This thematic area describes how the organizations conduct a comprehensive review of data \& evaluate the AI software before they routinely use the software in clinical practice. After the evaluation, an organizations decides to clinically integrate or abandon the AI software. Many organizations described the need for prospective validations of software in a ‘silent trial’ or ‘shadow mode.’ While prospective model validation takes place once the model is integrated into IT infrastructure, it is important to note that all other safety, effectiveness, and equity concerns should be taken into consideration throughout the entire process of AI adoption. Representative quotations are presented in Appendix \ref{appendix:quotes-5}.

\subsubsection{Data quality assurance}
The participants reported that assessing the quality of data used to train and validate the model was an important step for ensuring the safety and the efficacy of the model. Given that healthcare data is messy and may contain bias, it was important to perform thorough data quality assurance, even though it required significant resources. Some of the common practices of data quality assurance were ensuring that there was minimal bias present in data, creating a diverse data set for training the model, and cleaning data before building the model. 

\subsubsection{Validation}
Validating the model on real-time data prior to clinical integration was seen as a key component of assessing the safety, efficacy, and equity concerns of the model. Common methods of prospective validation included conducting a silent trial, comparing the model outcomes to patient charts in the medical record, and conducting a pilot on a subset of patients. The participants noted that model validation could be a time-consuming process, which could take many  months. Yet, it was an essential process to go through, especially when testing model accuracy on retrospective labels might not be sufficient for model validation due to inaccuracies in ground truth labels. The participants suggested that for successful model validation, algorithm developers should collaborate closely with clinical partners. Many also recommended validating the model on underserved populations as well as a large representative sample of populations to ensure equal model performance across different groups of patients.

\subsubsection{Risk mitigation}
To ensure the safe, effective, and equitable use of AI software, the participants indicated that organizations should be aware of the potential risks and have plans to mitigate them from early on. They suggested that being aware of the potential risks would be the first step. Cybersecurity risk, bias and disparities in care, and unexpected harms to patients were the most prevalent risks that the participants identified.

To mitigate these risks,  participants suggested that organizations should deeply understand regulatory requirements and work with a group of people with expertise in the full breadth of relevant domains when evaluating the model. For instance, the governance body should work closely with frontline workers who are knowledgeable and experienced in clinical care and community members to promote equity in healthcare. The participants pointed out that business administrators who do not have sufficient knowledge in patient care were typically the ones who evaluated AI algorithms. To ensure the safe, effective, and equitable use of AI software, participants emphasized that clinical personnel must be part of the evaluation process.

From a technical standpoint, the participants proposed to set clear eligibility criteria and model performance thresholds for a given model. Eligibility criteria defined which patient populations 
would be eligible for model use. If there were any disparities detected in the model performance among different patient populations, participants suggested addressing these issues by adjusting thresholds based on the disparities in the model performance. Then, once the model was ready for clinical integration, participants proposed to implement it incrementally in clinical settings to minimize the potential harm in patient care that could be caused by unforeseen safety issues.

\subsubsection{Clinically integrate vs. abandon decision}

After evaluating the safety, efficacy, and equity concerns of the model, the organizations decide to move forward with adopting the AI software or to stop the process. When all interested parties were satisfied with the model, they decided to integrate the model into clinical practice. However, when they were not satisfied with the model and did not perceive the benefit of adopting the model, they decided to abandon it. The participants reported that organizations often decide not to move forward with the adoption process when the model was no longer valuable, not implementable, not interoperable, too expensive, or unsuccessful in a pilot launch. The participants suggested that a decision to clinically integrate or abandon the model should be made based on evidence supported by data and that the organizations should be open to making an abandonment decision at any given point. One of the most prevalent challenges in making this decision was a lack of standardization in the decision-making process. The participants shared that having a standardized structure or guideline for risk assessment and clinical integration decisions would be helpful for making informed decisions.

\subsection{Execute AI product rollout, workflow integration, communication, education, and scaling}

This thematic area describes how the organizations provide necessary communication, education, and support to frontline workers expected to adopt the AI software, once they decide to integrate the model into clinical care. Representative quotations are presented in Appendix \ref{appendix:quotes-6}.

\subsubsection{Communication}
Communicating the plan for rollout of the tool was important for  successful clinical integration.
The plan for rollout explained when the model would be implemented in the workflow, who the end users would be, how to use the model, how to receive support and help in using the tool, and how to provide feedback on the tool. 
Bi-directional communication between developers and users was seen as key, both to allow adaptations needed to accommodate end users’ needs on the ground and to build a sense of accountability that AI developers  were listening to end users and respecting their agency. In communications leading up to clinical integration, it was also critical to explain the short-term and long-term benefits of using the tool and to answer potential questions from end users. 
 
The participants described that the launch of the tool was disseminated via various communication platforms, including email and website, depending on the target audience. Participants also reported that the communication was found to be more effective when it was made by clinical champions or trustworthy leaders in the organizations. The target audience of the communication included not only end users of the tool but also those who would oversee change management and those who would be impacted by the change.

\subsubsection{Education and training}
Participants described the need to tailor education and training to the target audience, their role, and the scale of change management. Education and training materials were disseminated in various ways, frequently including online learning management systems and webinars. One participant suggested that simulation-based training would be the most effective method of education and training.
 
Participants agreed that many operational and clinical parties have heard of AI but do not have much knowledge in it. Given the low AI literacy among front-line workers, participants wondered whether education and training should also offer foundational knowledge of AI.

\subsubsection{Trust}
Building an appropriate scope of trust between the AI software and end users was known to be a big challenge. Even if end users trusted the technology, there was the additional challenge of ensuring this trust did not extend beyond the model’s competence. Participants were aware of the concern that end-users would show automation bias and over-rely on the model. Participants hoped to build trust in AI software to a degree where clinical end users trusted the technology and found it useful, but not to a degree where they would fully rely on it in making decisions. 
 
The participants suggested a few approaches to build trust with end users. The first approach was to make algorithm developers have conversations with end users to provide transparent explanations about the model with trustworthy data and make them become familiar with AI and its use. If communicating directly with the end users was not feasible, the developers should at least speak with frontline clinicians or clinical leaders who were trusted by the end users. Gaining trust from them was found to be an efficient way to slowly build trust with the end user community. The second approach was to show end users how clinically valuable it would be to use the tool by showing them real-world evidence from the local context. One participant, for example, described an incident where they showed end users how the tool would be able to efficiently triage their patients. Another approach was to give end users agency and engage them from early stages of the AI software adoption process. That way, they could have transparency over the process, raise questions, provide feedback, and ultimately feel more confident about adopting the tool.

\subsection{After operationalization, monitor and maintain the AI product and impacted ecosystem}

This thematic area describes how the organizations monitor the technical and operational aspects of the AI software to ensure that it continues to function appropriately, if they decide to sustain the use of it. In this stage, we use the term impacted ecosystem to describe the environment in which front-line workers and patients are affected by use of the AI in practice. Representative quotations are presented in Appendix \ref{appendix:quotes-7}.

\subsubsection{Technical monitoring}

Technical monitoring involves monitoring the technical aspects of the model, such as model performance metrics, to ensure that the model continues to perform as intended. The participants cautioned that model evaluation may have to change over time, as the model performance may drift due to changes made in clinical care. They suggested that if the model drifts, the entire team of technical, clinical, and operational parties should get together to potentially retune the model and continue monitoring and revalidating the model over time.

\subsubsection{Operational monitoring}
Another type of monitoring that the participants described was operational monitoring. The participants reported that the biggest challenge in operational monitoring was staff turnover. Staff turnover, which happens frequently in clinical care, made the fidelity of model adoption particularly challenging, because it required the need for ongoing education and training. Additionally, when new tools became available, old tools tended to be abandoned and not used. The participants described how difficult it was to make clinicians continue to use existing tools when newly launched tools could take priority. To mitigate these challenges, the participants suggested creating robust documentation standards and governance rubrics for every tool clinically integrated. Participants also emphasized the importance of  creating new roles dedicated to overseeing operational monitoring.

\subsubsection{Outcome monitoring}
Outcomes were measured across various domains, including patient health outcomes, equity in patient care, financial outcomes, and staff satisfaction with the tool. 
Monitoring outcomes could be challenging in general, especially when there was a lack of infrastructure for quality monitoring.
Participants described the difficulty measuring the true impact of the AI component of a solution in isolation, because algorithms are often combined with treatment protocols, follow-up care, and changes in care-team responsibilities. This was particularly problematic when a model had been implemented for quite some time, as outcome measures defined prior to the model’s launch became less relevant benchmarks for the evaluation of current outcomes. 

Despite the challenges, outcome monitoring was still perceived to be important. 
For example, monitoring patient outcomes could address a limitation of technical monitoring when model performance might not be a reliable measure for patients receiving appropriate care. 
Monitoring staff satisfaction with the tool would inform why end users might use or not use the tool and identify areas of improvement.

\subsubsection{Cadence of monitoring reviews}
The participants reported that while it would not be practical to constantly have clinicians monitor model performance, periodic review of AI software monitoring and its use was important.
The participants emphasized that additional monitoring might be required when any updates or unexpected events (e.g. implementation of new medical equipment within the relevant workflow) took place in clinical settings. They noted that frequent monitoring would require significant resources, as more models were put into practice. To make the monitoring process more feasible, the participants suggested the creation of channels for users where they could report any concerns related to the models and potential adverse events that happen in clinical care. The frequency of AI system monitoring varied widely across organizations and use cases, ranging from 24/7 to once every 3 years.

\subsubsection{Accountability and ownership}
The participants reported that there was no clear ownership of lifecycle management activities. Participants often recommended a centralized group with dedicated resources be accountable for monitoring the AI software. They proposed that joint accountability between clinical, statistical, and technical providers would be ideal for continued use of the model. One challenge that they raised was how to engage clinicians, particularly end users, as product owners, as they were often stretched between too many responsibilities already.

\subsection{Update or decommission the AI product and impacted ecosystem}
This thematic area describes how organizations make updates to or decommission the AI software. Many participants shared that they had not yet updated or decommissioned a model. These participants reported their aspirations for a future state of organizational governance. Representative quotations are presented in Appendix \ref{appendix:quotes-8}.

\subsubsection{Update}
The participants reported that they would consider updating the model when changes in clinical care, intervention outcome, and IT environment were detected. They appreciated that often a model cannot be simply retrained, because the prior clinical integration changed the environment. Participants also recommended conducting user studies and quantitative analyses to understand potential problems to address before deciding on model updates.

\subsubsection{Decommission}
Participants reported that they would consider decommissioning a model if the model performed poorly, when safety concerns emerged, when the model would be replaced by a newly available better model, when there was no value in the clinical use case, and when the model was no longer financially supported. A rare example of model decommissioning was observed with software used for capacity management during the early stages of COVID-19. Some participants reported that they had to decommission software that was developed during the peak of the COVID-19 pandemic because they no longer found it to be useful.
 
The participants indicated that decisions  to decommission a model are very challenging, especially when there is no clear ownership or delegation of responsibility defined for monitoring the model. However, they noted that when a serious safety issue is identified, even with little notice, a decision to decommission the model should be made urgently.

\section{Discussion}
\label{sec:discussion}

In this project, we conducted usability-testing sessions to design scaffolding for AI adoption standards that health system leaders found immediately useful in practice. Concurrently, we conducted interviews to examine the current and aspirational state of AI adoption in healthcare. The earlier design process informed how the latter qualitative research results could be framed around key decision points. By identifying the major challenges to AI adoption in healthcare more broadly, we were positioned to surface key decision points and community needs to work towards a future shared standard for AI adoption. In the end, our qualitative research revealed 8 key decision points across the entire process of adopting AI. Taken together, our findings show how health systems currently make decisions to adopt AI into clinical practice and are in need of shared approaches to governing the process. To our knowledge, this project is the first multi-organizational collaboration with leaders from a range of major health systems across the United States to investigate organizational governance processes of AI adoption. 

Overall, we found that health systems are developing individual institutional practices and principles to guide their process of AI adoption but acknowledge an urgent need for a move towards shared standards. 
Not only does the absence of shared frameworks create social and economic costs on individual health systems, but it also impedes adoption of AI and undermines safety, efficacy, and equity. 
For example, vendors have a vested interest in positioning their products as not requiring regulatory approval or fitting under discretionary enforcement. This regulatory uncertainty 
externalizes costs onto health systems that would be otherwise borne by the vendor. This burden shifting is wildly inefficient as health systems assessing the same AI products must engage in duplicate work and less well-resourced health systems will be further disadvantaged in their struggle to invest in adequate assessment and validation capabilities. Furthermore, even better resourced health systems will eventually shift away from engaging with external vendors, in order to prioritize internally developed AI products because it allows them to better assess and manage risk.

The uncertain regulatory posture also chills learning. Our research revealed that uncertainty about best practices, stemming in part from the regulatory void, makes health systems more reluctant to be transparent with external interested parties, including other health systems and vendors, about the ways AI products are used and governed in practice. This chill on information exchange is deeply problematic, given the need for developing standardized governance processes to fill the regulatory gap.

While there is no explicitly named shared and comprehensive set of best practices and processes for AI adoption in healthcare, our interviews surfaced a set of common practices that health systems do already engage in. The interviews highlighted the importance of testing, validating, and monitoring clinical care delivery and its outcomes, rather than the AI performance. The focus on the health care delivery ecosystem aligns with the growing emphasis on the understanding that a machine learning model is part of a socio-technical system, and that the other components of the system need to be modeled \cite{selbst2019fairness}. Health systems should consider both the humans (whose experience will shift due to the introduction of the AI), the policy environment (comprising regulation, professional obligations and organizational policy) and the practical considerations of clinical care to improve measured outcomes.

Our interviews also revealed agreement on the need for iterative evaluation with interdisciplinary teams that include clinicians, technical experts, and data scientists. Interdisciplinary collaborations promote efficient assessment of requirements, implementation feasibility, and development of outcome measures. They also ensure the quality of clinical care by addressing different scenarios of use case, potential risks in bias, security, privacy, and mitigations. Specifically, our interviewees underscored the need for evaluation and integration methods that foster calibrated clinician trust, while emphasizing that AI must be viewed as a trustworthy but not infallible source of information. Developing documentation and explanatory artifacts of the AI model and data may build clinician knowledge throughout the integration process. Respecting clinician’s agency and expertise may reduce unwarranted skepticism about AI outputs. 

Finally, as indicated earlier, our interviews reported that AI governance requires new resources and expertise beyond current capabilities within health systems that align with the categories of algorithmic curators, brokers, and articulators \cite{kellogg2020algorithms}. Algorithmic curators standardize data and make them interoperable across different settings to properly  validate algorithms. Algorithmic brokers communicate the value of AI systems to clinicians and other affected parties. 
Algorithmic brokers working for AI vendors, however, were viewed as playing a more pernicious role. Their access to health system leaders could subvert the problem-led and principle guided processes around AI adoption and integration. Socio-legal scholars similarly warn that governance in private hands can produce symbolic structures that legitimate the behavior of regulated entities without furthering public policy objectives \cite{edelman1992legal}. Algorithmic articulators, including clinicians, data scientists, and technologists, are responsible for change management which is necessary to improve the quality of clinical integration. Interviewees described that the introduction of AI systems may require the creation of new teams to interact with or monitor the system, new communication strategies to maintain professional boundaries, and new expertise. This articulation work was viewed as essential to success but difficult to fully identify prior to clinical integration. The interviews also shed light on the inequities that may arise from the absence and uneven distribution of resources and expertise across business units within a single institution and across institutions themselves.

\section{Conclusion}
\label{sec:conclusion}

Our research identified an immediate opportunity to improve organizational governance of AI in healthcare. The current research identifies missing resources and suggests that a range of new professionals and practices are considered necessary to produce knowledge that informs the actions taken by clinicians. Governance processes are required at the organizational, not algorithmic, level for analyzing the quality of AI and clinical outcomes produced by socio-technical systems reconfigured by AI. Achieving a healthy governance ecosystem requires a process to align and expand the competencies of the existing actors-regulators and standard setting bodies–in a manner that best furthers the policy goals of safety, efficacy and equity. Successfully governing AI requires a detailed understanding of the competencies and capacities of different actors. As an important building block to produce organizational governance, each health system must develop governance frameworks along with assigning responsibilities to entities with the independence, expertise, and operational capacity. Our research identified 8 key decision points shared among health systems that can serve as an initial use case for developing shared standards. Health system organizations must exchange information, foster dialogue, and contribute best practices that complement and strengthen a rapidly evolving regulatory landscape.

\begin{acks}
We thank the Gordon and Betty Moore Foundation for supporting the project. We thank all interview participants for sharing their insights and time with us. We thank Joanne Kim and Claire Carroll for conducting preliminary work of the project, and Shira Zilberstein for contributing to the later stage of the project. We thank Duke Heart Center for helping administer and manage the grant. 

\end{acks}

\clearpage

\bibliographystyle{ACM-Reference-Format}
\bibliography{main}

\clearpage

\appendix

\clearpage

\section{Antimicrobial Stewardship}
\label{appendix:antimicrobial}

\begin{table}[h!]
  \begin{center}
    \captionsetup{justification=raggedright,singlelinecheck=false}
    \caption{Journey of antimicrobial stewardship.}
    \label{tab:antimicrobial}
    \begin{tabular}{|m{3cm}|m{13cm}|}
        \hline
         Year & Events \\ \hline
1928 &  
Alexander Fleming conducted a series of experiments that led to the discovery of penicillin, the world’s first antibiotic \cite{bennett2001alexander}. 
\\ \hline
1939 - 1945 &
Penicillin became mass produced as a medication during World War II.
\\ \hline
1945 &
 In response to the identification of penicillin-resistant bacteria, Fleming warned against the potential harms of inappropriate penicillin use in The New York Times: “The microbes are educated to resist penicillin and a host of penicillin-fast organisms is bred out... In such cases the thoughtless person playing with penicillin is morally responsible for the death of the man who finally succumbs to infection with the penicillin-resistant organism. I hope this evil can be averted." By 1999, most specimens of certain bacteria isolated from intensive care units in the United States were penicillin resistant \cite{fishman2012policy}.
\\ \hline
1945&
Fleming won the Nobel Prize in Medicine, and penicillin was attributed with saving over 200 million lives.
\\ \hline
1997&
Professional societies of physicians were the first interested party to address antibiotic resistance in a coordinated fashion. The Society for Healthcare Epidemiology of America (SHEA) and the Infectious Diseases Society of America (IDSA) published guidelines for the prevention of antimicrobial resistance in hospitals \cite{shlaes1997society}.
\\ \hline
2007 & 
The Society for Healthcare Epidemiology of America (SHEA) and the Infectious Diseases Society of America (IDSA) updated guidelines \cite{dellit2007infectious}.
\\ \hline
2008&
Legislative action ensuring implementation of antimicrobial stewardship programs followed thereafter. California became the first state to require all hospitals to develop processes for evaluating the judicious use of antibiotics \cite{trivedi2013state}.
\\ \hline
2014&
The Centers for Disease Control and Prevention (CDC) released guidelines for antimicrobial stewardship programs \cite{pollack2014core}. 
\\ \hline
2017&
Antimicrobial stewardship programs became required for any hospital seeking Joint Commission accreditation \cite{baker2019leading}.
\\ \hline
2019&
Federal regulation passed requiring hospitals to establish antimicrobial stewardship programs in order to receive the Centers for Medicare and Medicaid Services (CMS) payments \cite{centers2014medicare}.
\\ \hline

  \end{tabular}
  \end{center}
\end{table}

\clearpage

\section{Design Insights}
\label{appendix:ideo-insights}

\begin{table}[h]
  \begin{center}
    \caption{Major design insights around experiences and needs.}
    \label{tab:ideo}
    \begin{tabular}{|l|l|}
        \hline
         Insight & Descriptions \\ \hline
Birds of a feather, & Championing the adoption of a technology-enabled solution involves many  \\
 not yet flying together & stakeholders. Yet, the community feels out of reach. \\ \hline
Dancing mostly  & There’s a craving for standardizing procurement, anchored on external benchmarks.  \\
to the same tune & Yet, every institution has their own unique resources and challenges. \\ \hline
The squeaky wheel  & Without shared, objective frameworks for adopting AI solutions, individual  \\
gets the grease & incentives of stakeholders in the procurement process outweigh decisions that  \\
& optimize for safety, efficacy, and equity. \\ \hline
Treading lightly & When the conditions for open, candid conversations about safety, performance, and  \\
& bias in technology occur, stakeholders lack confidence in ways to implement actions \\
&  that mitigate harm. \\ \hline
Three steps forward,  & AI is adopted with the goal of benefiting clinicians. Yet, clinical care staff are not ready  \\
two steps back & to engage with and incorporate AI’s recommendations, which leads to increased  \\
& inefficiencies and workload. \\ \hline
Bringing the cart and the horse & When tech solution adoption is not nestled within change management efforts,  \\
& solutions fail to deliver on their potential because the workflow challenges have yet \\
&  to be solved. \\ \hline
Learn about us from us & In centering the clinician as the customer, patients and community are absent from  \\
& decisions with the potential to build or break trust in the care they receive. \\ \hline
    \end{tabular}
  \end{center}
\end{table}

\clearpage

\begin{figure}[b!]
    \begin{center}
    \caption{Prototype structured around mock key decision points.}
    \includegraphics[width=0.59\textwidth]{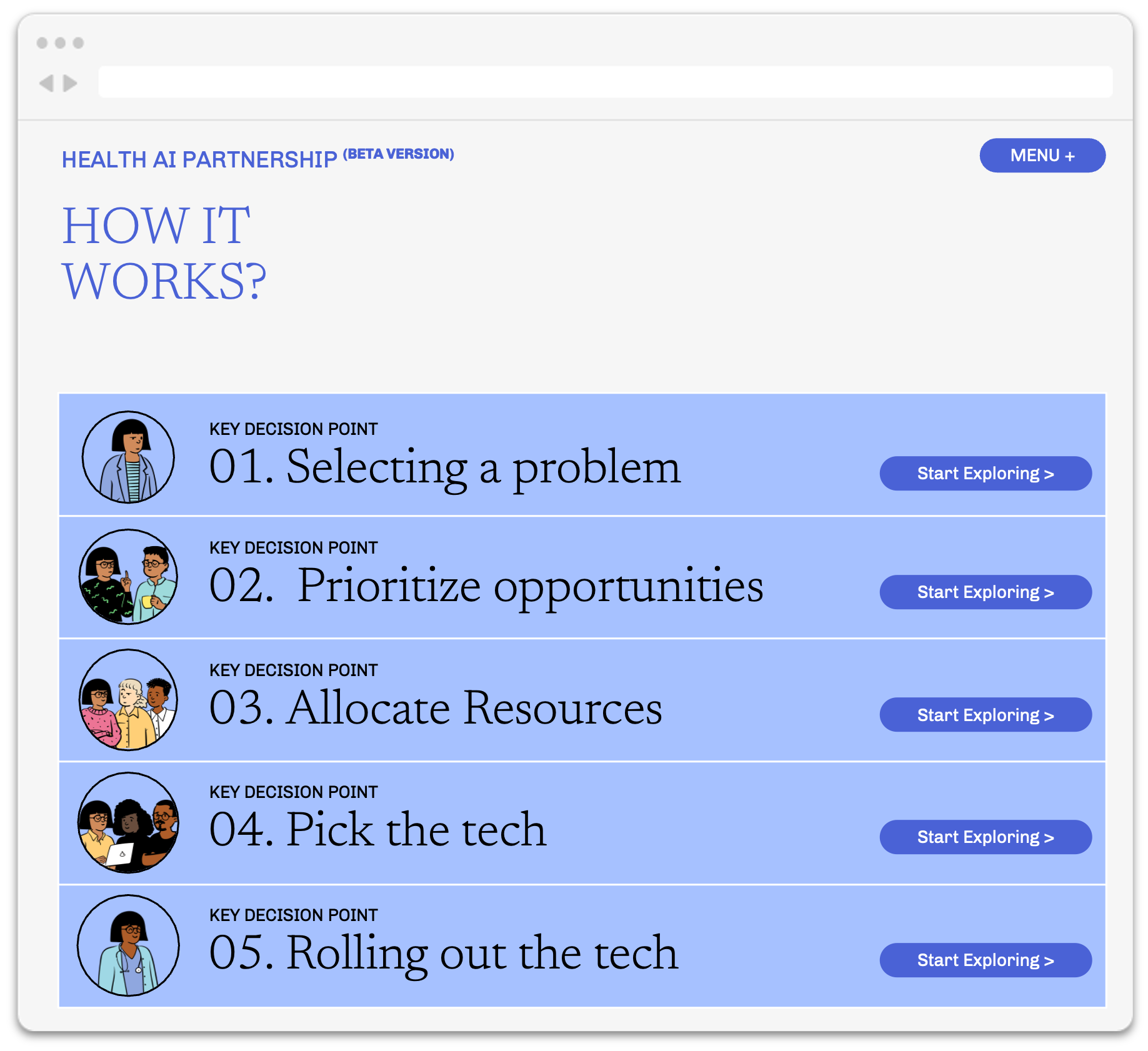}
    \label{fig:ideo-deliverable}
    \end{center}
\end{figure}

\clearpage

\section{Materials for Interview Guide}
\label{appendix:four-stages}


\begin{table}[h!]
  \begin{center}
    \captionsetup{justification=raggedright,singlelinecheck=false}
    \caption{Four stages of AI adoption.}
    \label{tab:four-stages}
    \begin{tabular}{|m{3cm}|m{13cm}|}
        \hline
         Stage & Descriptions \\ \hline
Problem identification and procurement &  The first stage starts with identifying a pain point and ends with allocating a  budget to pursue an opportunity.  \\ \hline
Development \;\;\;\;\;\;\;\;\;\;\;\;\;\; and adaptation & The second stage starts with developing an algorithm internally or adapting an externally developed one for internal use and ends with a decision to use the algorithm in patient care.   \\ \hline
Clinical integration & The third stage starts with a rollout of the algorithm, including the associated  training, communication, and change management, and ends with a decision  to make the algorithm use standard operations.   \\ \hline
Lifecycle management & The fourth stage starts with the algorithm used in steady state and ends with a decision to either continue using the algorithm, update the algorithm, or  decommission the algorithm.  \\ \hline
    \end{tabular}
  \end{center}
\end{table}

\begin{table}
\label{tab: interview}
  \begin{center}
    \captionsetup{justification=raggedright,singlelinecheck=false}
    \caption{Sample interview questions.}
    \label{tab:interview}
    \begin{tabular}{|m{3cm}|m{13cm}|}
        \hline
         Stage & Questions \\ \hline
Problem identification and procurement &  
\begin{itemize}
    \item Can you describe the process by which [organization or organizational unit] selects AI projects to invest in?
    \item Can you describe the process for evaluating AI opportunities? How does it differ from other software? 
    \item How does [organization or organizational unit] approach the “build versus buy” decision for AI tools? 
    \item When deciding to invest in an AI project, what stakeholders does [organization or organizational unit] seek input from?

\end{itemize}
\\ \hline
Development \;\;\;\;\;\;\;\;\;\;\;\;\;\; and adaptation &  
\begin{itemize}
    \item How is [organization or organizational unit] building capabilities to develop new AI tools?
    \item How does [organization or organizational unit] assess the safety and efficacy of AI tools? Are there specific metrics you look for?
    \item Can you describe how new workflows or roles are designed to optimize use of new AI tools? 
    \item What is the decision-making process to move forward with integrating an AI tool into clinical care? Are there specific metrics you look for? 
    \item What stakeholders are involved in the design of the AI tool? And the development of the AI tool?
\end{itemize}
\\ \hline
Clinical integration &  
\begin{itemize}
    \item Can you describe the process by which [organization or organizational unit] integrates new AI tools into clinical workflows? 
    \item What is the process for scaling a new AI tool into new settings?
    \item Who are the stakeholders involved in the roll out of new AI tool? Who helps with communication, education, and change management?
\end{itemize}
\\ \hline
Lifecycle management &  
\begin{itemize}
    \item How does [organization or organizational unit] monitor the use of newly adopted technologies? How does this differ for AI?
    \item Under what circumstances does [organization or organizational unit] decide to decommission a technology used in clinical practice? How does this differ for AI?
    \item How do clinicians and staff in [organization or organizational unit] get feedback on their use and impact of new AI tools?
    \item Who is responsible for the lifecycle management of new AI tools? 

\end{itemize}
\\ \hline

    \end{tabular}
  \end{center}
\end{table}


\clearpage

\section{Decision Point 1 Quotes}
\label{appendix:quotes-1}

\begin{table}[h!]
\label{tab:decision-points-1}
  \begin{center}
  \captionsetup{justification=raggedright,singlelinecheck=false}
    \caption{Representative quotes from Decision Point 1.}
    \label{tab:quotes-1}
    \begin{tabular}{|m{3cm}|m{13cm}|}
        \hline
         Subtheme & Quotes \\ \hline
Problem identification &  
\begin{itemize}
    \item ``What I mean by the push and pull is the pull is that [organization] executives and clinician often identify areas where they believe would benefit from advanced analytics and predictive models and they bring those to the forefront. … The other part, it's a bit of the push, where we identify areas that we believe would be important to emerging opportunities to invest time to develop ahead of a need being articulated.'' [Operational Role]
    \item ``It happens at the C-suite [executive-managers within an organization]. It happens in the frontlines. So, we'll have situations where faculty will have a connection with a company and they'll also have some leadership role in their department or division. And they will bring a new technology, like an AI technology, into their group for evaluation. And sometimes that happens, and we only find out about it later that it hasn't gone through the appropriate risk assessment, security, assessment, etc.'' [Technical Role]
\end{itemize}
\\ \hline
Problem prioritization &  
\begin{itemize}
    \item ``Harm, heat and heart. Harm is quantitative evidence that we're doing something poorly like maybe our mortality in some subset is higher than expected than the state norm or than similar places. Heat would be … a problem which we need to have a stance on and be working on. And then, heart is just the stories that pull at the heartstrings.'' [Technical Role]
    \item ``Compiling an inventory, a catalog of AI capabilities that we have already in house. And what that will do is allow us to identify opportunities, where we can leverage what we already have, and extend one of our current capabilities to cover something that is new or just the gap. And then, once we do that, as the first stage, we will prioritize through that committee.'' [Operational Role]
\end{itemize}
\\ \hline
AI contribution &  
\begin{itemize}
    \item ``Right now, a lot of the AI solutions are basically trying to do the same thing as a doctor. So, it's like, take an x-ray, read it as a radiologist would. But we already have radiologists, so what is this thing doing? I think the very promising applications are basically triage. … We can use it to triage the easy cases from the hard cases so that you can then reallocate radiologists’ effort in useful ways to the hard cases. So, I think finding that right interface between the human and the algorithm seems super important.'' [Bias Key Informant]
    \item ``I see these products are really being able to sharpen our clinical predictive skills. So much of our time, and effort and moral distress is spent with clinical uncertainty and a lot of money. So, if we can be better at predicting which patients would do well, which won't, which patients will benefit from surgery, which won't. … Where there's greatest clinical uncertainty, we're using a data set in order to make better predictions. That's where I see the greatest utility.'' [Technical Role]
\end{itemize}
\\ \hline

    \end{tabular}
  \end{center}
\end{table}

\clearpage

\section{Decision Point 2 Quotes}
\label{appendix:quotes-2}

\begin{table}[h!]
\label{tab:decision-points-2}
  \begin{center}  \captionsetup{justification=raggedright,singlelinecheck=false}
    \caption{Representative quotes from Decision Point 2.}
    \label{tab:quotes-2}
    \begin{tabular}{|m{3cm}|m{13cm}|}
        \hline
         Subtheme & Quotes \\ \hline
Requirement \;\;\;\;\;\;\;\;\;\;\;\;\;\; and feasibility &  
\begin{itemize}
    \item ``I'd say you need finance; you need a data analyst; you need a software engineer to do the integration; you need your IT system to be involved. And then you need some clinicians, or a physician champion, or a clinic nurse champion. And then you need who's going to be following up post and managing the relationship. And then you need to think about the contracts that you develop.'' [Bias Key Informant]
    \item ``We have lots of people asking us: 'This is my idea. Can you please confirm [that regulatory approvals are not required]?' or 'This is my idea. If you don't think that [it] is going to fall into some of these exceptions or enforcement discretion, what can we do to fall on that side of things?' That's definitely an area of lots and lots of interest, because there's such a big jump in resources required, time, cost, if you are on the other side of the line, and you are regulated.'' [Regulation Key Informant]
\end{itemize}
\\ \hline
Quality assurance &  
\begin{itemize}
    \item ``Big consideration was a similar workflow for managing patients. ... We were thinking in a bigger picture trying to minimize the number of vendors that were partnering with us on this, trying to minimize the number of systems that we have to keep track of in the overall picture of radiology IT and minimizing the number of interfaces between systems, because we will obviously want some of these results to go to PACS (Picture Archive and Communication System) for clinical radiologists to review during the context of patient care. So, in the interest of trying to minimize these things, we weren't just looking for a partner that could provide good stroke AI offerings, but who could also partner with us on potentially … commercializing our internally developed tools, but also somebody who would be interested in expanding beyond neuro AI to do other imaging, AI tools and other body areas as well.'' [Clinical Role]
    \item ``I don't think we even really have a great understanding of how to measure an algorithm’s performance, let alone its performance across different race and ethnic groups. … If we're assessing like a drug, we understand that we want to know the treatment effects for a given outcome. And we know we have to do a randomized trial to get there. I don't think that's exactly what we need to do for algorithms. But there does need to be some infrastructure for defining what we mean by this algorithm works and assessing whether it works as well for group A and group B.'' [Bias Key Informant]
\end{itemize}

\\ \hline
Build vs. buy decision &  
\begin{itemize}
    \item ``If you have the sample size, you will tend to do better if you have the expertise to develop it on your own in your own setting. The reason being that, you're not working with a black box algorithm, so you can actually see what you're working with.'' [Bias Key Informant]
    \item ``Even trying to use external tools … has been a challenge from a regulatory perspective. There's still a lot of concern about sending data on the cloud through when you send GPUs (Graphics Processing Unit).'' [Technical Role]
\end{itemize}
\\ \hline
Aligning \;\;\;\;\;\;\;\;\;\;\;\;\;\;\;\;\;\;\;\;\;\;\;\;\;\;\; affected parties &  
\begin{itemize}
    \item ``Should an ethics person be involved in those sorts of discussions? I mean, I think that would be helpful probably partly because maybe it would be helpful to have somebody who has a kind of 10,000-foot perspective about the many different ways in which a system could fail that individual stakeholders might miss in their more like parochial perspectives.'' [Ethics Key Informant]
    \item ``We're trying to drive standardization … because I believe that we would benefit from it. It's a question of alignment of groups and how they cooperate and alignment of priority.'' [Technical Role]
\end{itemize}
\\ \hline

    \end{tabular}
  \end{center}
\end{table}

\clearpage

\section{Decision Point 3 Quotes}
\label{appendix:quotes-3}

\begin{table}[h!]
\label{tab:decision-points-3}
  \begin{center}
  \captionsetup{justification=raggedright,singlelinecheck=false}
    \caption{Representative quotes from Decision Point 3.}
    \label{tab:quotes-3}
    \begin{tabular}{|m{3cm}|m{13cm}|}
        \hline
         Subtheme & Quotes  \\ \hline
Model performance measures &  
\begin{itemize}
    \item ``I think, a couple of metrics that are useful at the point of internal validation or preliminary external validation is looking at the unidirectional and threshold specific performance measures, which are very simple arithmetic. It's like a false positive rate, false negative rate at the clinical threshold that you think you're actually going to use. And false negative rate in particular, I think people get very concerned that it doesn't have certain statistical properties, but in practice, it really closely measures the lived reality of patients that you're interested in, when you're thinking about equity. It's like, who needs something, and who isn't getting it, is a concrete thing.'' [Bias Key Informant]
    \item ``I think coming up with a relatively simple test of evaluating outputs, stratified by different variables that you think maybe of concern, either doing that yourself or asking the vendor to do it, that feels like very threshold to me. I think if you have very low resources, that's probably the one thing you should do.'' [Operational Role]
\end{itemize}
\\ \hline
Success measures &  
\begin{itemize}
    \item ``I always encourage them to use those four aims which would be reducing cost, improving patient care, the patient experience, and then reducing burden on the part of healthcare providers. So where are they? And what are the metrics that you would capture if this solution were successful? Just to engage, does it make sense for us to pursue it?'' [Clinical Role]
    \item ``We've been pretty good at saying … this is what we want to try to accomplish. And that typically is expressed by the clinicians in one way in terms of patient affecting things—could be performance measurements, could be quality of life, could be lots of different things—and then it's expressed by the business side in a different way. Again, reduced rehospitalization or reduced resource utilization, things that are, if you will, measurable.'' [Clinical Role]
\end{itemize}
\\ \hline

  \end{tabular}
  \end{center}
\end{table}

\clearpage

\section{Decision Point 4 Quotes}
\label{appendix:quotes-4}

\begin{table}[h!]
\label{tab:decision-points-4}
  \begin{center}  \captionsetup{justification=raggedright,singlelinecheck=false}
    \caption{Representative quotes from Decision Point 4.}
    \label{tab:quotes-4}
    \begin{tabular}{|m{3cm}|m{13cm}|}
        \hline
         Subtheme & Quotes \\ \hline
Operational \;\;\;\;\;\;\;\;\;\;\;\;\;\; optimization &  
\begin{itemize}
    \item ``We reached out to the nurses that would be using the tool itself and asked their opinions of how, where to put the monitors, what would be easiest for them. How can we implement this so that it made it as simple as possible so that it wasn't a burden to use the tool?'' [Operational Role]
    \item ``EHRs (Electronic Health Records), our lab has done some work on an intelligent autocomplete for filling out triage notes in the ED (Emergency Department). And when I first heard about that, I thought this is a no brainer, like clinicians are gonna love having autocomplete as you understand it. It's not been as popular as you would expect. And it's not because the algorithm is wrong. The algorithm is pretty spot on. But it doesn't fit in their workflow.'' [Bias Key Informant]
\end{itemize}
\\ \hline
Technical optimization &  
\begin{itemize}
    \item ``If we can simplify the output of the model down to a single number or a single color code or something that is put in one small place of an [EHR], some [EHR] user interface element, that's what we should do. We don't want to over complicate things.'' [Technical Role]
    \item ``If you're building a model, and you say, ‘We can predict with incredible accuracy and early on with these vital signs. We just need that cognitive decline measure.’ And the nurses sometimes put it in, but you know what, they don't always. So, let's put a BPA (Best Practice Alert) in so that we get them to enter the cognitive decline every shift because they're not always doing it. It's like no, no, that's just not ideal because you're impacting the workflow.'' [Clinical Role]
\end{itemize}
\\ \hline

  \end{tabular}
  \end{center}
\end{table}

\clearpage

\section{Decision Point 5 Quotes}
\label{appendix:quotes-5}

\begin{table}[h!]
\label{tab:decision-points-5}
  \begin{center} \captionsetup{justification=raggedright,singlelinecheck=false}
    \caption{Representative quotes from Decision Point 5.}
    \label{tab:quotes-5}
    \begin{tabular}{|m{3cm}|m{13cm}|}
        \hline
         Subtheme & Quotes \\ \hline
Data quality assurance &  
\begin{itemize}
    \item ``This is to create very diverse data sets. And the way we do that is not convenience sampling but is really like pulling every data set and then using stratified sampling. The pooling of every data set takes a lot of time.'' [Bias Key Informant]
    \item ``The real worry is far more subtle that there are patterns in society that produce patterns in data. Patterns of social exclusion, lack of access, that produces patterns in data where patients look sicker at baseline. And we're kind of like, yeah, that's normal.'' [Ethics Key Informant]
\end{itemize}
\\ \hline
Validation &  
\begin{itemize}
    \item ``For the clinical validation piece, we work with our close clinical partners. … The data scientists have stories about why some features might look relevant and others not. The clinician kind of gets involved with that as well. And then once the lead clinician feels good about the model, then we do some interrater reliability test typically with some physicians. … And they kind of go through how well do they agree with each other? How well do they agree with the model? It's always a nice feeling when they agree with each other and the model at the same rate. Everyone feels good about that. There is an art to that because there's no like, defined line there. But what we're really working towards is, are we positioning them [clinical team] to be the authority on what's deployed in a pilot.'' [Operational Role]
    \item ``The biggest problem is the way we evaluate algorithms is still accuracy, and accuracy based on what we have now in the data. And, of course, we know that that shouldn't be our goal post because it's such a flawed ground truth to be evaluating our algorithms against.'' [Bias Key Informant]
\end{itemize}
\\ \hline
Risk mitigation &  
\begin{itemize}
    \item ``You can actually work with more diverse groups that aren't necessarily technology based—ones that may be knowledgeable of human centered design—and have them work on how the technology is actually implemented and the impacts of that. I think having a broad mix of people that work within that area will at least give you an idea of some of the impacts of the technology when that bias is discovered because it will come up. The question is just when and how soon are you able to identify it and mitigate the negative impacts?'' [Community Key Informant]
    \item ``Is there a rules-based approach policy by which we determine what algorithms to run on me, as opposed to which ones to run on you, as opposed to which ones we run on mom? Do we reinforce it? Are there scenarios where we basically know to exclude?'' [Technical Role]
\end{itemize}
\\ \hline
Clinically integrate vs. abandon decision &
\begin{itemize}
    \item ``He had all the data and our response, or our amount of calls prior to implementing and post, the percentage of patients that were actually rapids [patients requiring urgent review] that were scoring higher. So, he had a lot of data that he took to the decision meeting.'' [Clinical Role]
    \item ``We've gone into silent mode and proven that the model is not worth going into full production, meaning it costs too much, is squeezed, isn't worth the rub. The intervention isn't going to be set up, and we've not gone to production.'' [Operational Role]
     \item ``I don't think the vast majority of systems that click the box to turn this on run it through a compliance or technical governance infrastructure or something like that. They just said, ‘This is a new capacity that [EHR] is offering. Sounds like it's useful. Let's go for it.'' [Regulation Key Informant]
\end{itemize}
\\ \hline

  \end{tabular}
  \end{center}
\end{table}

\clearpage

\section{Decision Point 6 Quotes}
\label{appendix:quotes-6}

\begin{table}[h!]
\label{tab:decision-points-6}
  \begin{center} \captionsetup{justification=raggedright,singlelinecheck=false}
    \caption{Representative quotes from Decision Point 6.}
    \label{tab:quotes-6}
    \begin{tabular}{|m{3cm}|m{13cm}|}
        \hline
         Subtheme & Quotes  \\ \hline
Communication &  
\begin{itemize}
    \item ``So, trying to make sure that we're both doing kind of awareness campaigns as broad marketing that we then have email or web-based dissemination, and then that we have a meeting based approach to [promote] awareness that something is launching and [that we are] looking for feedback, and then iterating based on feedback after a pilot approach.'' [Clinical Role]
    \item ``It really helps to talk to providers, or whoever is going to benefit from this tool. Often, they're not aware. So, we rolled something out. … There was a lot of communication. But then, when we start monitoring some of those, the usage of those tools for features, people are not using them. … A lot of these communication and re-training, and just make sure that they are aware of the benefits of this tool and what is the long-term impact.'' [Technical Role]
\end{itemize}
\\ \hline
Education and training &  
\begin{itemize}
    \item ``Ideally, what I would like is—this may not be possible for resource constrained ideas—you have the clinician sit down with the interface and have them practice using it on a curated data set. … Maybe you got 100 clinicians in a room or whatever. And you do that walk through with 10 cases.'' [Bias Key Informant]
    \item ``Education. Straight, straight education, making sure that those individuals understand what the tool is about, what it's doing, what its inputs and outputs are, how it's evaluated. … It's going to be the toughest. And that's because a lot of these organizations, including cases across our company, may not understand what AI is. They've heard of AI, they've heard it will make, you know, decisions based on input, input information. But they really don't understand it.'' [Technical Role]
\end{itemize}
\\ \hline
Trust &  
\begin{itemize}
    \item ``One of the things that we found is not only giving them the predictive score is important, but they want to know what is driving that. So, we have … built out a tool for a visualization report. And it basically outlines what the top five predictors are. … If there's a BPA trigger, when it triggers and why it triggered? So, giving them the actual clinical information pulled from the EHR that shows what is contributing to that high-risk score. That is what we've done. That's actually helped fill in those gaps.'' [Operational Role]
    \item ``If we have a situation in which the machine is basically all the time right, doctors are just going to trust it and stop focusing on it. If we have a system where the system is wrong, lots of the time doctors aren't going to use it. If we have a system, on the other hand, where the system is wrong enough that doctors should be checking it a decent amount, and they find that they're fixing it up a decent amount, it's right there in that sweet spot. It's hard for me to imagine that it's staying there in the sweet spot, or frankly, that's a good use of physician time.'' [Regulation Key Informant]
\end{itemize}
\\ \hline

  \end{tabular}
  \end{center}
\end{table}

\clearpage

\section{Decision Point 7 Quotes}
\label{appendix:quotes-7}

\begin{table}[h!]
\label{tab:decision-points-7}
  \begin{center} \captionsetup{justification=raggedright,singlelinecheck=false}
    \caption{Representative quotes from Decision Point 7.}
    \label{tab:quotes-7}
    \begin{tabular}{|m{2cm}|m{15cm}|}
        \hline
         Subtheme & Quotes  \\ \hline
Technical \;\;\; monitoring &  
\begin{itemize}
    \item ``As you know, with AI and ML (Machine Learning), it's going to change over time. So, we're gonna have to keep revalidating that the model is not drifting. So, I think we're gonna get the data scientists involved. So, you're gonna have to keep doing those analyses in a different way than simply overriding or not, or whether it's still firing accurately or not, if that makes sense, because the definition of accurate may change.'' [Technical Role]
    \item ``I think one major thing to worry about is a distribution shift. I think there's been sort of abundant evidence that this is an issue in clinical settings. And it's necessarily not even just related to demographics. You change the format of your ICD (International Classification Diseases) codes, and now you're screwed, or you move between x-ray machines, just very banal pedestrian stuff like this.'' [Bias Key Informant]
\end{itemize}
\\ \hline
Operational monitoring &  
\begin{itemize}
    \item ``I think containerizing everything so that if someone were to leave, we would have the container to keep all the code, the operating environments to at least sort of have the bandage between transitions. So, I think it kind of boils down to how much we document and how we keep the good technical principles.'' [Technical Role]
    \item ``Because other tools come up. And then they take priority. … There's also turnover, while these clinics have a lot of turnovers. So, trying to keep new staff educated on what's available, what's out there, it's a constant challenge. I know [organization] has implemented a lot of tools and reports other things over the years that don't get used, because members often just don't know they're there, for whatever reason, because there's so much that's there now. They often just follow the workflows they need to follow and aren’t looking for anything new to do because they can barely keep up with what they have to do. This is an ongoing challenge.'' [Technical Role]
\end{itemize}
\\ \hline
Outcome \;\;\; monitoring &  
\begin{itemize}
    \item ``If the model is actually working, then the patient should get intervention, and then that intervention should change the outcome. So, we are not planning to monitor in terms of accuracy but going to the referrals and long-term depression outcomes, because we felt like accuracy won't be accurately measured as a practical validation.'' [Technical Role]
    \item ``I think most health systems are pretty shitty at figuring out how well it actually turns out in individual patient cases and that's part of the reason we don't have anything close to a learning health system, because we're not good at monitoring outcomes, except in a few weird, unusual cases. And that's a problem. … I don't think it's that different from the infrastructure that should be in place for quality monitoring in health systems more generally.'' [Regulation Key Informant]
\end{itemize}
\\ \hline
Cadence of \;\;\;\;\; monitoring \;\;\;\;\; reviews &
\begin{itemize}
    \item ``And then there needs to be quarterly reassessments where you have every algorithm and the metrics we've all agreed are the right metrics. How is it doing? Is it still doing well? Or has there been some weird shift?'' [Bias Key Informant]
    \item ``I think our users are not going to be good at figuring out what's going wrong. They all see that the computer is telling them something funky. So, I do think that when it comes to providing clinical IT support, having an infrastructure that is 24/7 is key, and it's expensive, and it's hard to build out.'' [Technical Role]
    \end{itemize}
\\ \hline
Accountability and ownership &
\begin{itemize}
     \item ``I would say it should be a triangle. I think there's three responsible parties that should be continuing to maintain models: the clinical provider [the clinician owner of the model], the technical provider [the model developer or vendor]. But the third piece that I think is missing is the statistical support. Because it's impossible to ask those other two arms to say that the model is still performing accurately, if you're not sort of consistently looking at the predictive characteristics of the model.'' [Clinical Role]
     \item ``Who runs this? What should they be asking about? How do we systemize this? It probably shouldn’t be the clinician that I work with who’s been in charge of the whole project. She has a full-time job, you know what I mean?'' [Bias Key Informant]
\end{itemize}
\\ \hline

  \end{tabular}
  \end{center}
\end{table}

\clearpage

\section{Decision Point 8 Quotes}
\label{appendix:quotes-8}

\begin{table}[h!]
\label{tab:decision-points-8}
  \begin{center}
  \captionsetup{justification=raggedright,singlelinecheck=false}
    \caption{Representative quotes from Decision Point 8.}
    \label{tab:quotes-8}
    \begin{tabular}{|m{3cm}|m{13cm}|}
        \hline
         Subtheme & Quotes  \\ \hline
Update &  
\begin{itemize}
    \item ``Let's say the algorithm does seem to be getting good performance, clinicians aren't complaining that much about it, but empirically, you see their usage drops off a bit. Then that's maybe like, let's do a user study. We can keep using the algorithm, but we need to talk to half a dozen people about whether they're feeling alarm fatigue, or how we can make the interface more user friendly. It's hard for me to think of a situation where I'm like, ‘oh, just retrain the algorithm without doing any further analysis.'" [Bias Key Informant]
    \item ``Retrain the model—I would argue that you can't do that. Because you are not dealing with the same system that you started with before the deployment of the model. What's happened is that there was a health system, there were people doing their thing [in the] absence of a model, they were basically trying to predict themselves, whatever they were doing. Deployed a model. The model itself changes healthcare, just changes the system.'' [Technical Role]
\end{itemize}
\\ \hline
Decommission &  
\begin{itemize}
    \item ``It's not kind of moving anything positive for our patients, and we get negative feedback from our clinical teams. Those would be the kind of cases where we'd say, ‘okay, this is not actually relevant.'" [Operational Role]
    \item ``It is really hard to decommission things. Really, really hard.'' [Operational Role]
\end{itemize}
\\ \hline

  \end{tabular}
  \end{center}
\end{table}


\end{document}